\pdfoutput=1

\documentclass[11pt]{article}

\usepackage{acl}

\usepackage{times}
\usepackage{latexsym}
\usepackage{amsmath}
\usepackage{url}
\usepackage{bbding}
\usepackage{colortbl}
\usepackage{graphicx}
\usepackage{makecell}
\usepackage{algorithm}
\usepackage{algpseudocode}
\usepackage{multicol}

\renewcommand{\Return}[1]{\State{\textbf{return} #1}}
\newcounter{mytempeqncnt}


\usepackage[T1]{fontenc}

\usepackage[utf8]{inputenc}

\usepackage{microtype}

\title{Entailment Graph Learning with Textual Entailment and Soft Transitivity}

\author{Zhibin Chen$^{123}$\quad Yansong Feng$^{13}$\thanks{\ \ Corresponding author.}\quad Dongyan Zhao$^{123}$\\
$^1$Wangxuan Institute of Computer Technology, Peking University, China\\
$^2$Center for Data Science, Peking University, China\\
$^3$The MOE Key Laboratory of Computational Linguistics, Peking University, China\\
\texttt{\{czb-peking, fengyansong, zhaody\}@pku.edu.cn}}

\begin{document}
\maketitle
\begin{abstract}
Typed entailment graphs try to learn the entailment relations between predicates from text and model them as edges between predicate nodes. The construction of entailment graphs usually suffers from severe sparsity and unreliability of distributional similarity. We propose a two-stage method, Entailment Graph with Textual Entailment and Transitivity (EGT2). EGT2 learns local entailment relations by recognizing possible textual entailment between template sentences formed by typed CCG-parsed predicates. Based on the generated local graph, EGT2 then uses three novel soft transitivity constraints to consider the logical transitivity in entailment structures. Experiments on benchmark datasets show that EGT2 can well model the transitivity in entailment graph to alleviate the sparsity issue, and lead to significant improvement over current state-of-the-art methods\footnote{Our code is available at \url{https://github.com/ZacharyChenpk/EGT2}.}.
\end{abstract}

\section{Introduction}

Entailment, as an important relation in natural language processing (NLP), is critical to semantic understanding and natural language inference (NLI)
. Entailment relation has been widely applied in different NLP tasks such as Question Answering~\citep{pathak2021scientific,khot2018scitail}, Machine Translation~\citep{pado2009measuring} and Knowledge Graph Completion~\citep{yoshikawa2019combining}. When coming across a question that \emph{"Which medicine cures the infection?"}, one can recognize the information \emph{"Griseofulvin is preferred for the infection,"} in the corpus and appropriately write down the answer with the knowledge that \emph{"is preferred for"} entails \emph{"cures"} when their arguments are \textit{medicines} and \textit{diseases}, although the surface form of predicate \emph{"cures"} does not exactly appear in the corpus. There are many ways to present one question, and it is impossible to handle them without understanding the entailment relations behind the predicates. Previous works on analyzing entailment mainly focus on Recognizing Textual Entailment (RTE) between pairs of sentences, and many recent attempts have achieved quite promising performance in detecting entailment relations using transformer-based language models~\citep{he2020deberta, JMLR:v21:20-074, schmitt-schutze-2021-language}.

\label{intro}

\begin{figure}[t!]
    \centering
    \includegraphics[width=0.95\linewidth]{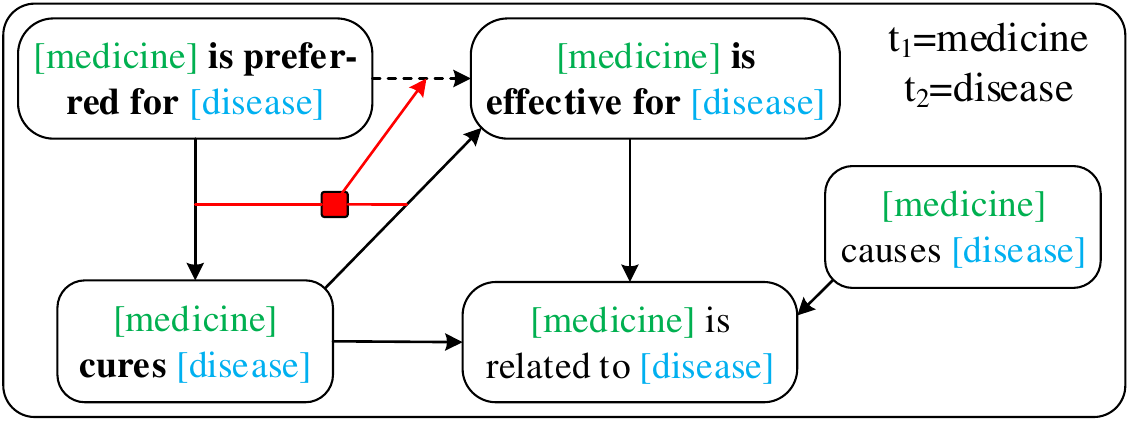}
    \caption{A simple example of entailment graph with types \emph{medicine} and \emph{disease}. The dashed line represents a missing entailment recovered by considering the transitivity constraint (red) based on the two premise entailment between three \textbf{boldfaced} predicates.
    }
    \label{fig:example}
\end{figure}

By modeling typed predicates as nodes and entailment relations as directed edges, the \textbf{Entailment Graph} (\textbf{EG}) is a powerful and well-established form to represent the context-independent entailment relations between predicates and reflect the global features of entailment inference, such as paraphrasing and transitivity. As EGs are able to help reasoning without additional context or resource, they can be seen as a special type of structural knowledge in natural language. Figure~\ref{fig:example} shows an excerpt entailment graph about two types of arguments, \emph{Medicine} and \emph{Disease}. Generally speaking, an entailment graphs can be built based on a three-step process: \textbf{extracting} predicate pairs from a corpus, building \textbf{local} graphs with locally computed entailment scores, and modifying the graphs with \textbf{global} methods.

However, existing EG construction methods still face challenges in both local and global stages. The Distributional Inclusion Hypothesis (DIH) about entailment assumes that given a predicate (relation) $p$, it can be replaced in any context by another predicate (relation) $q$ if and only if $p$ entails $q$ \citep{geffet-dagan-2005-distributional}. Most local methods in previous works are guided by DIH, thus rely on the distributional co-occurrences from corpora, including named entities, entity pairs and context, as features to compute the \textit{local} entailment scores. Since different predicate pairs are processed independently, the locally built graphs suffer from severe \textbf{data sparsity}. That is, there are many entailment relations missing (as edges) in the graphs if the predicate pairs do not co-occur in the corpus. Furthermore, predictions from local models may  not be coherent with each other, for example, a local model may output three predictions like, \textit{$a$ entails $b$},  \textit{$b$ entails $c$} and \textit{$c$ entails $a$} at the same time, which actually indicate possible errors among the local predictions.

To overcome the challenges faced by local models, different global approaches are used to take the interactions and dependencies between entailment relations into consideration. The first discussed global dependency is the logical \emph{transitivity} among different predicates, that is, predicate $a$ entails predicate $c$ if there is another predicate $b$ making both "$a$ entails $b$" and "$b$ entails $c$" hold simultaneously. \citet{berant-etal-2011-global} uses the Integer Linear Programming (ILP) to ensure the transitivity constraints on the entailment graphs, which is , unfortunately, not scalable on large graphs with thousands of nodes. \citet{hosseini-etal-2018-learning} models the structural similarity across graphs and paraphrasing relations within graphs to learn the global consistency, but does not gain further improvement due to the lack of \textbf{high-quality local graphs} and proper \textbf{transitivity} modeling.

In order to deal with the problems in the local and global stages, we propose a novel entailment graph learning approach, \textbf{E}ntailment \textbf{G}raph with \textbf{T}extual Entailment and \textbf{T}ransitivity (\textbf{EGT2}). EGT2 builds high-quality local entailment graphs by inputting predicates as sentences into a transformer-based language model fine-tuned on an RTE task to avoid the unreliability of distributional scores, and models the global transitivity on these scores through carefully designed soft constraint losses, which alleviate the data sparsity and are feasible on large-scale local graphs.
Our key insight is that the entailment relation $a\to c$ correctly implied by the transitivity constraint is based on two conditions: (1) the appropriate constraint scalable on large graphs containing rich information, and (2) the reliability of local graphs offering the premise $a\to b$ and $b\to c$, which is impractical for previous distributional approaches, but may be available for the models well-behaved on RTE tasks. Specifically, the input sentences fed to transformer-based language models are formed without context, which makes our method accessible to those predicates not appearing in the corpus. The transitivity implication is confined to entailment relations with high confidence, which improves the quality of implied edges and cuts down the computational overheads. In a word, this paper makes the following contributions:
\begin{itemize}
    \item we present a novel approach based on textual entailment to scoring predicate pairs on local entailment graphs, which is reliable without distributional features and valid for arbitrary predicate pairs.
    \item we present three carefully designed global soft constraint loss functions to model the transitivity among entailment relations on large entailment graphs, thus alleviate the data sparsity issue of previous local approaches.
    \item we evaluate our method on benchmark datasets, and show that our EGT2 significantly outperforms previous entailment graphs construction approaches. The further analysis proves that our local and global approaches are both useful for learning entailment graphs.
\end{itemize}
\section{Related Work}

\label{related-work}

Based on DIH, previous works extract feature vectors for typed predicates to compute the local distributional similarity. The set of entity argument pair strings, like \emph{"Griseofulvin-infection"} in the example of Section~\ref{intro}, are used as the features weighted by Pointwise Mutual Information \citep{berant-etal-2015-efficient, hosseini-etal-2018-learning}. Given the feature vectors for a  predicate pair, different similarity scores, like cosine similarity, Lin \citep{lin-1998-automatic-retrieval}, DIRT \citep{lin2001discovery}, Weeds \citep{weeds-weir-2003-general} and Balanced Inclusion \citep{szpektor-dagan-2008-learning}, are calculated as the local similarities. \citet{hosseini-etal-2019-duality} and \citet{hosseiniopen} use Markov Chain on an entity-predicate bipartite graph weighted by link prediction scores to calculate the transition probability between two predicates as the local score. They rely on the link predication model to generate the features in fact. \citet{guillou-etal-2020-incorporating} adds temporal information into entailment graphs 
by extracting entity pairs within a limited 
temporal window as predicate features. \citet{mckenna2021multivalent} extends the graphs to include entailment relations between predicates with different numbers of arguments by splitting the features from argument pairs into independent entity slots, which impairs the representation ability of features when unary predicates are involved.

As mentioned in Section \ref{intro}, entailment graphs are generally learned by imposing global constraints on the local entailment relations about extracted predicates. The transitivity in entailment graphs is modeled by the Integer Linear Programming (ILP) in \citet{berant-etal-2011-global}, which selects a transitive sub-graph of a local weighted graph to maximize the summation over the weights of its edges. Their work is limited to a few hundreds of predicates due to the computational complexity of ILP. For better scalability, \citet{berant-etal-2012-efficient} and \citet{ berant-etal-2015-efficient} make a strong FRG-assumption that \textit{if predicate $a$ entails predicates $b$ and $c$, $b$ and $c$ entail each other}, and an approximation method, called Tree-Node-Fix (TNF). Obviously, the assumption is too strong to be satisfied by real cases. 

Since the hard constraints are difficult to work well on large-scale entailment graphs, \citet{hosseini-etal-2018-learning} propose two global soft constraints that maintain the similarity between paraphrasing predicates within typed graphs and between predicates with the same names in graphs with different argument types. Their soft constraints are also used in \citet{hosseini-etal-2019-duality} and \citet{hosseiniopen}. The similarity between paraphrasing predicates, which ensures $(a\to c)\odot(b\to c)$ and $(c\to a)\odot(c\to b)$ when $a\leftrightarrow b$, implicitly takes the transitivity between paraphrasing predicates and third predicate into consideration. But it ignores the transitivity in more common cases, and leads to a limited improvement on performance.

Meanwhile, the transformer-based Language Model (LM), although proved to be effective in RTE tasks~\citep{he2020deberta, JMLR:v21:20-074, schmitt-schutze-2021-language}, has received less attention in entailment graph learning. \citet{schmitt2021continuous} uses pretrained LM on the Lexical Inference in Context (LIiC) task, which is closely related to entailment graph learning. \citet{hosseiniopen} uses pretrained BERT to initialize the contextualized embeddings in their contextualized link prediction and entailment score calculation. Higher scores are assigned to the entailed predicates in the context of their premises, which is one implicit expression form of DIH and different from our direct utilization of LM on textual entailment.
\begin{table*}
\caption{Examples of sentence generator $S$.}\label{example-sent}
\centering
\quad
\begin{tabular}{l|l}
\hline
\textbf{Predicates} & \textbf{Sentences} \\
\hline
(be.1,be.capital.of.2,location$_1$,location$_2$) & Location A is capital of Location B.\\
(contain.1,contain.2,location$_2$,location$_1$) & Location B contains Location A.\\
(prefer.2,prefer.for.2,medicine,disease) & Medicine A is preferred for Disease B.\\
(give.2,give.3,person,thing) & Person A is given Thing B.\\
(aggrieved.by.2,aggrieved.felt.1,thing,person) & Person B feels aggrieved by Thing A.\\
\hline
\end{tabular}
\end{table*}

\section{Our Method: EGT2}

\subsection{Definition and Notations}

The goal of entailment graph learning is to extract predicates, learn the entailment relations and build entailment graphs from raw text corpora. Following previous works \citep{hosseini-etal-2018-learning, hosseini-etal-2019-duality}, we use the binary relations from neo-Davisonian semantics as predicates, which is a type of first-order logic with event identifiers. For instance, with the semantic parser (here, GraphParser \citep{reddy-etal-2014-large}), the sentence: \centerline{\emph{"Griseofulvin is preferred for the infection."}} can be transformed into the logical form 
\centerline{$\exists e.prefer_2(e,Griseofulvin)$}
\centerline{$\cap prefer_{for}(e,infection)$}
where $e$ denotes an event. By considering a relation for each pair of extracted arguments, this sentence refers to one predicate, $p=$ \emph{(prefer.2,prefer.for.2,medicine,disease)} \footnote{The numbers after the predicate words are corresponding argument positions of entity \emph{"Griseofulvin"} (second argument of \emph{prefer}) and \emph{"infection"} (second argument of the preposition \emph{for}), and the later two items are the types of arguments.}. Likely, the sentence \emph{"Griseofulvin cures the infection."} contains $q=$ \emph{(cure.1,cure.2,medicine,disease)}. Formally, a predicate with argument types $t_1$ and $t_2$ is represented as $p=(w_{p,1}.i_{p,1}, w_{p,2}.i_{p,2}, t_1 ,t_2)$. The event-based predicate form is strong enough to describe most of the relations in real cases \citep{parsons1990events}.

With $T$ as the set of types and $P$ as the set of all typed predicates, $V(t_1,t_2)$ contains typed predicates $p$ with unordered argument types $t_1$ and $t_2$, where $p\in P$ and $t_1,t_2\in T$. For predicate $p=(w_{p,1}.i_{p,1}, w_{p,2}.i_{p,2}, t_1 ,t_2)$, we denote that $\tau_1(p)=t_1$, $\tau_2(p)=t_2$ and $\pi(p)=(w_{p,1}.i_{p,1}, w_{p,2}.i_{p,2})$. In other words, $V(t_1,t_2)=\{p|(\tau_1(p)=t_1\land\tau_2(p)=t_2)\lor(\tau_1(p)=t_2\land\tau_2(p)=t_1)\}$. 

A typed entailment graph $G(t_1,t_2)=<V(t_1,t_2), E(t_1,t_2)>$ is composed of the nodes of typed predicates $V(t_1,t_2)$ and the weighted edges $E(t_1,t_2)$. The edges can be also represented as sparse score matrix $W(t_1,t_2)\in [0,1]^{|V(t_1,t_2)|\times|V(t_1,t_2)|}$, containing the entailment scores between predicates with type $t_1$ and $t_2$. As the different argument types can naturally determine whether two predicates have the same order of arguments, the order of argument type is not important while $t_1\ne t_2$, and therefore we can ensure that $G(t_1,t_2)=G(t_2,t_1)$. For those predicates $p$ with $\tau_1(p)=\tau_2(p)$, the two argument types are labeled with orders, which allows the graph to contain the entailment relations with different argument orders, like \emph{(be.1,be.capital.of.2,location$_1$,location$_2$)} $\to$ \emph{(contain.1,contain.2,location$_2$,location$_1$)}.

\subsection{Local Entailment based on Textual Entailment}

Inspired by the outstanding performance of pretrained and fine-tuned LMs on RTE task, which is closely related to the entailment graphs, EGT2 uses fine-tuned transformer-based LM to calculate the local entailment scores of typed predicated pairs. 

In order to utilize the knowledge about entailment relations in pretrained and fine-tuned LM, EGT2 firstly transfers the predicate pair $(p,q)$ into corresponding sentence pair $(S(p),S(q))$ by sentence generator $S$, as the complicated predicates cannot be directly input into the LM. For typed predicate $p=(w_{p,1}.i_{p,1}, w_{p,2}.i_{p,2}, t_1 ,t_2)$, the generator deduces the positions of arguments about the predicate based on $i_{p,1}$ and $i_{p,2}$, generates the surface form of $p$ based on $w_{p,1}$ and $w_{p,2}$, and finally concatenates the surface form with capitalized types as its arguments. Some generated examples are shown in Table \ref{example-sent}, and the detailed algorithm of $S$ is described in Appendix \ref{sec:sg}.

After generating sentence pair $(S(p),S(q))$ for predicate pair $(p,q)$, EGT2 inputs $(S(p),S(q))$ into a transformer-based LM to calculate the probability of the entailment relation $p\to q$ as the local entailment score in $G(t_1,t_2)$. In our experiments, the LM is implemented as DeBERTa \citep{he2020deberta}. Generally, an entailment-oriented LM will output three scores for a sentence pair, representing the probability of relationship \emph{entail}, \emph{contradict} and \emph{neutral} respectively. Formally, we denote the weighted matrix of local entailment graph with type $t_1$ and $t_2$ as $W^{local}$, and the weight of the edge between $p$ and $q$ in $W^{local}$ is calculated as:

\begin{equation}
\begin{aligned}
\label{localbuild}
    W^{local}_{p,q} &= P(p\to q)\in [0,1],\\
    P(p\to q)&=\frac{e^{LM(\emph{entail}|p,q)}}{\sum_{r\in \{\emph{entail,contradict,neutral}\}} e^{LM(r|p,q)}},
\end{aligned}
\end{equation}
where $LM(r|p,q)$ is the output score of corresponding relationship by the LM. As the local entailment is based on the LM fine-tuned to perform textual entailment, the local graph can be built for any predicates in the parsed semantic form, or in any other forms by changing sentence generator $S$.

\subsection{Global Entailment with Soft Transitivity Constraint}

\begin{figure*}[!t]
\normalsize
\setcounter{mytempeqncnt}{\value{equation}}
\setcounter{equation}{1}
\begin{equation}
\begin{aligned}
    \label{loss}
    L_1 &= -log\prod_{a,b,c\in V(t_1,t_2),\atop W_{a,b},W_{b,c}>1-\epsilon}\min(1,\frac{W_{a,c}}{W_{a,b}W_{b,c}})\\
    &= \sum_{a,b,c\in V(t_1,t_2)}I_{1-\epsilon}(W_{a,b})I_{1-\epsilon}(W_{b,c})ReLU(logW_{a,b}+logW_{b,c}-logW_{a,c})\\
    L_2 &= \sum_{a,b,c\in V(t_1,t_2)}-I_{1-\epsilon}(W_{a,b})I_{1-\epsilon}(W_{b,c})I_0(W_{a,b}W_{b,c}-W_{a,c})logW_{a,c}\\
    L_3 &= \sum_{a,b,c\in V(t_1,t_2)}-I_{1-\epsilon}(W_{a,b})I_{1-\epsilon}(W_{b,c})I_0(W_{a,b}W_{b,c}-W_{a,c})W_{a,b}W_{b,c}logW_{a,c}\
\end{aligned}
\end{equation}
\setcounter{equation}{\value{mytempeqncnt}}
\setcounter{equation}{2}
\vspace{-3.5mm}
\end{figure*}

Existing approaches use global learning to find correct entailment relations which are missing or underestimated in local entailment graphs to overcome the data sparsity. Following \citet{hosseini-etal-2018-learning}, the evidence from existing local edges with high confidence is used by EGT2 to predict missing edges in the entailment graphs. 

The \emph{transitivity} in entailment relation inference implies $a\to c$ while both $a\to b$ and $b\to c$ hold. For instance, in the example of Figure \ref{fig:example}, the entailment \emph{"is preferred for" $\to$ "is effective for"} is discovered because \emph{"is preferred for" $\to$ "cures"} and \emph{"cures" $\to$ "is effective for"} have been learned. The key challenge to incorporate the transitivity constraint into weighted graphs is discreteness of logical rules. Discreteness makes the rules impossible to be directly used in gradient-based learning methods without NP-hard complexity, as different predicate pairs are jointly involved in the calculation. To unify the discrete logical rules with gradient-based learning, inspired by \citet{li-etal-2019-logic}, EGT2 uses the logical constraints in the form of differentiable triangular norms \citep{gupta1991theory, klement2013triangular}, or called t-norms, as the \textbf{soft constraints} so that the gradient-based learning methods can be applied. 

Different t-norm methods transfer the discrete rules into different continuous loss functions. Traditional product t-norm maps $P(A\land B)$ into $P(A)P(B)$, $P(A\lor B)$ into $P(A)+P(B)-P(A)P(B)$, and $P(A\to B)$ into $\min(1,\frac{P(B)}{P(A)})$. For the entailment relations, the probability of transitivity to be satisfied is:
\setcounter{equation}{2}
\begin{equation}
\begin{aligned}
    &P[(a\to b \land b \to c)\to(a\to c)]\\
    =&\min(1,\frac{W_{a,c}}{W_{a,b}W_{b,c}}),
\end{aligned}
\end{equation}
where the probability of the entailment relation $a\to b$ is represented by the local entailment scores $W_{a,b}$. To alleviate the noise from those edges assigned low confidence by local LM, EGT2 only takes the local edges whose scores are higher than $1-\epsilon$ into account (as $a\to b$ and $b\to c$), where $\epsilon$ is a small hyper-parameter because the local probability scores tend to be close to $0$ or $1$ in practice. Therefore, to maximize the probability of transitivity constraint satisfied over all predicates in the entailment graph $G(t_1,t_2)$, EGT2 tries to minimize the following minus-log-likelihood loss function $L_1$ in Eq. \ref{loss}, where $I_y(x)=1$ if $x>y$, or $0$ otherwise.

Another important t-norm, called the Gödel t-norm, maps $P(A\to B)$ into $1$ if $P(B)\ge P(A)$ or $P(B)$ otherwise. Therefore, the Gödel probability of transitivity to be satisfied is:

\setcounter{equation}{3}
\begin{equation}
\begin{aligned}
    &P[(a\to b \land b \to c)\to(a\to c)]\\
    =&\left\{
    \begin{array}{lr}
         W_{a,c}& W_{a,b}W_{b,c}>W_{a,c}\\
         1& otherwise
    \end{array}
    \right.,
\end{aligned}
\end{equation} and EGT2 similarly tries to minimize the loss function $L_2$ in Eq. \ref{loss}. It should be noted that transitivity constraints will be disobeyed not only by the missing edges, but also by the spurious edges in the local graphs. Therefore, we expect the soft constraints to take reducing the weights of premise edges into consideration. $L_1$ achieves this by the loss item $W_{a,b}$ and $W_{b,c}$, and we modify $L_2$ to $L_3$ in Eq. \ref{loss} so that the low confidence of $W_{a,c}$ will help to detect whether $W_{a,b}$ and $W_{b,c}$ are spurious. Our t-norm soft constraints, although do not guarantee the obedience of transitivity, are effective approximations for the transitivity property.

Given the local entailment graph $G(t_1,t_2)$ with weighted edges $W^{local}$, in order to ensure that the global entailment graph $W$ is not too far from $W^{local}$, EGT2 finally minimizes the following loss function $L$ to trade off the distance from local graphs and the soft transitivity constraint:

\begin{equation}
\begin{aligned}
\label{finalloss}
    L = \sum_{a,b\in V}(W_{a,b}-W^{local}_{a,b})^2+\lambda L_i,\ i=1,2,3
\end{aligned}
\end{equation}
where $L_i$ is the specified implementation of soft transitivity constraint in Eq. \ref{loss}, and $\lambda$ is a non-negative hyper-parameter that controls the influence of two loss terms.

\section{Experimental Setup}

\subsection{Predicate Extraction}

Following \citet{hosseini-etal-2018-learning} and \citet{hosseini-etal-2019-duality}, we use the multiple-source NewsSpike corpus \citep{zhang-weld-2013-harvesting}, which contains 550K news articles, to extract binary relations as generated predicates in EGT2. We make use of the triples released and filtered in \citet{hosseini-etal-2019-duality}, which applies GraphParser \citep{reddy-etal-2014-large} based on Combinatorial Categorial Grammar (CCG) syntactic derivations to extracting binary relations between predicates and arguments. The argument entities are linked to Freebase \citep{bollacker2008freebase} and mapped to the first level of FIGER types \citep{ling2012fine} hierarchy. The type of a predicate is determined by its two corresponding argument entities. The triples are filtered by two rules to remove the noisy binary relations and arguments: (1) we only keep those argument-pairs appearing in at least 3 relations; (2) we only keep those relations with at least 3 different argument-pairs. The number of relations in the corpus is reduced from 26M to 3.9M, covering 304K typed predicates in 355 typed entailment graphs. Only those predicate pairs co-occurring with at least one same entity-pair (e.g., \emph{Griseofulvin-infection}) will be linked to calculate the local scores, and as a result, our local predicate pairs are identical with \citet{hosseini-etal-2019-duality}. As we focus on using global models to alleviate the sparsity of local edges, more potential methods to extracting denser local edges will be studied in our future research.

\subsection{Evaluation Datasets and Metrics}
\label{metrics}

We use Levy/Holt Dataset \citep{levy-dagan-2016-annotating, holt2019probabilistic} and Berant Dataset \citep{berant-etal-2011-global} to evaluate the performance of entailment graph models. 

In Levy's dataset, each example contains a pair of triples with the same entities but different predicates. Some questions with one predicate were shown to the annotating workers, like \emph{"Which medicine cures the infection?"}. The label for each example are either \emph{True} or \emph{False}, indicating whether the first typed predicate entails the second one, by asking the workers whether the first predicates can answer the question with the second one. For example, if \emph{"Griseofulvin is preferred for the infection"} is a correct answer of the above question, the dataset labels \emph{"is preferred for"} $\to$ \emph{"cures"}. \citet{holt2019probabilistic} re-annotates Levy's dataset and forms a new dataset with 18,407 examples (3,916 positive and 14,491 negative), referred as Levy/Holt Dataset. The dataset is split into validation set (30\%) and test set (70\%) as \citet{hosseini-etal-2018-learning} in our experiments.

\citet{berant-etal-2011-global} annotates all the entailment relations in their corpus, which generates 3,427 positive and 35,585 negative examples, referred as Berant Dataset. Their entity types do not exactly match with the first level of FIGER types hierarchy, and therefore a simple hand-mapping by \citet{hosseini-etal-2018-learning} is used to unify the predicate types.

To be comparable with previous works, we evaluate our methods on the test set of Levy/Holt Dataset and the whole Berant Dataset by calculating the area under the curves (AUC) with changing the classification threshold of global entailment scores. \citet{hosseini-etal-2018-learning} argues that the AUC of Precision-Recall Curve (PRC) for precisions in the range $[0.5,1]$, as predictions with higher precision than \emph{random} are more important for the downstream applications. Therefore, we report both the AUC of PRC
for precisions in the range $[0.5,1]$ and the traditional AUC of ROC, which is more widely used in evaluation of other tasks. 

\subsection{Comparison Methods}

We compare our model with existing entailment graph construction methods \citep{berant-etal-2011-global, hosseini-etal-2018-learning, hosseini-etal-2019-duality, hosseiniopen} and the best local distributional method, Balanced Inclusion \citep{szpektor-dagan-2008-learning}, referred as BInc. We also include ablation variants of our EGT2, including local models with or without fine-tuning. 

\subsection{Implementation Details}
\label{implement}

For local transformer-based LM, EGT2 uses DeBERTa \citep{he2020deberta} implemented by the Hugging Face transformers library \citep{wolf2019huggingface}\footnote{https://github.com/huggingface/transformers}, which has been fine-tuned on MNLI \citep{williams-etal-2018-broad} dataset. In order to adapt it to the special type-oriented sentence pattern generated by $S$, we expand the validation set by extracting all of the predicates, generating sentence pairs by generator $S$ for every two predicates, and checking whether they are labeled as paraphrase or entailment in the Paraphrase Database collection (PPDB) \citep{pavlick-etal-2015-ppdb}. We split $80\%$ of the generated corpus to fine-tune the DeBERTa with Cross-Entropy Loss, and the rest as the validation set of fine-tuning process. The fine-tuning learning rate $\alpha_f=10^{-5}$, and the process is terminated while the $F_1$ score of \emph{entail} on validation set does not increase in 10 epochs or training after 100 epochs. 

\begin{table}
\caption{Model performance on Levy/Holt Dataset and Berant Dataset.
The best performances on every metric are \textbf{boldfaced}. Results with $*$ are from the original papers.}\label{result-table}
\centering
\begin{tabular}{l|c|c|c|c}
\hline
\textbf{Methods} & \multicolumn{2}{c|}{Levy/Holt} & \multicolumn{2}{c}{Berant}\\
\hline
\textbf{Metrics} & PRC & ROC & PRC & ROC\\
\hline
BInc & .155 & .632 & .147 & .677\\
Local-Sup & .161 & .632 & .129 & .651\\
Hosseini18 & .163 & .637 & .174 & .682\\
Hosseini19$^*$ & .187 & - & - & - \\
 - Local & .167 & .639 & .118 & .378 \\
Hosseini21$^*$ & .195 & - & - & -\\
\hline
EGT2-Local & .313 & .712 & .360 & .857\\
- w/o Fine-tuning& .234 & .673 & .147 & .732 \\
EGT2-$L_1$ & .345 & \bf.761 & .437 & \bf.880\\
EGT2-$L_2$ & .319 & .755 & .361 & .879\\
EGT2-$L_3$ & \bf.356 & .755 & \bf.443 & .871\\
\hline
\end{tabular}
\end{table}

For global soft transitivity constrains, we use SGD \citep{cun1998efficient} to optimize the scores $W$ in entailment graphs with loss function $L$ in Eq. \ref{finalloss} for $e=5$ epochs. The SGD learning rate $\alpha=0.05$, the coefficient $\lambda=1$, and the confidence threshold $\epsilon=0.02$. The hyper-parameters are selected based on Levy/Holt validation dataset. More implementation details are given in Appendix \ref{addi-detail}.

For testing, if one or both predicates of the example do not appear in the corresponding typed entailment graph, we handle the example as untyped one by resorting to its average score among all typed entailment graphs. This setting is also used for all local and global methods in the experiments for fair comparison.
\section{Experiment Results and Discussion}

\subsection{Main Results}

We summarize the model performances on both Levy/Holt and Berant datasets in Table \ref{result-table}. All global methods, including \citet{hosseini-etal-2018-learning}, \citet{hosseini-etal-2019-duality} and EGT2, perform better than their corresponding local methods, which demonstrates the effect of global constraints in alleviating the data sparsity. Although using the same extracted entailment relations with \citet{hosseini-etal-2019-duality}, our EGT2-Local significantly outperforms previous local methods because of the high-quality entailment scores generated by reliable fine-tuned textual entailment LM. On the whole, 
EGT2 with transitivity constraint $L_3$ outperforms all the other models on both Levy/Holt Dataset and Berant Dataset with AUC of PRC, while EGT2-$L_1$ performs best with AUC of ROC. All of three soft transitivity constraints boost the performance of local model on all evaluation metrics, which shows that making use of transitivity rule between entailment relations improves the local entailment graph. EGT2-$L_1$ or EGT2-$L_3$ performs better than EGT2-$L_2$, which indicates that involving the premises $a\to b$ and $b\to c$ into loss function is also important for using transitivity constraints.

\begin{figure*}[t!]
    \centering
    \includegraphics[width=0.95\linewidth]{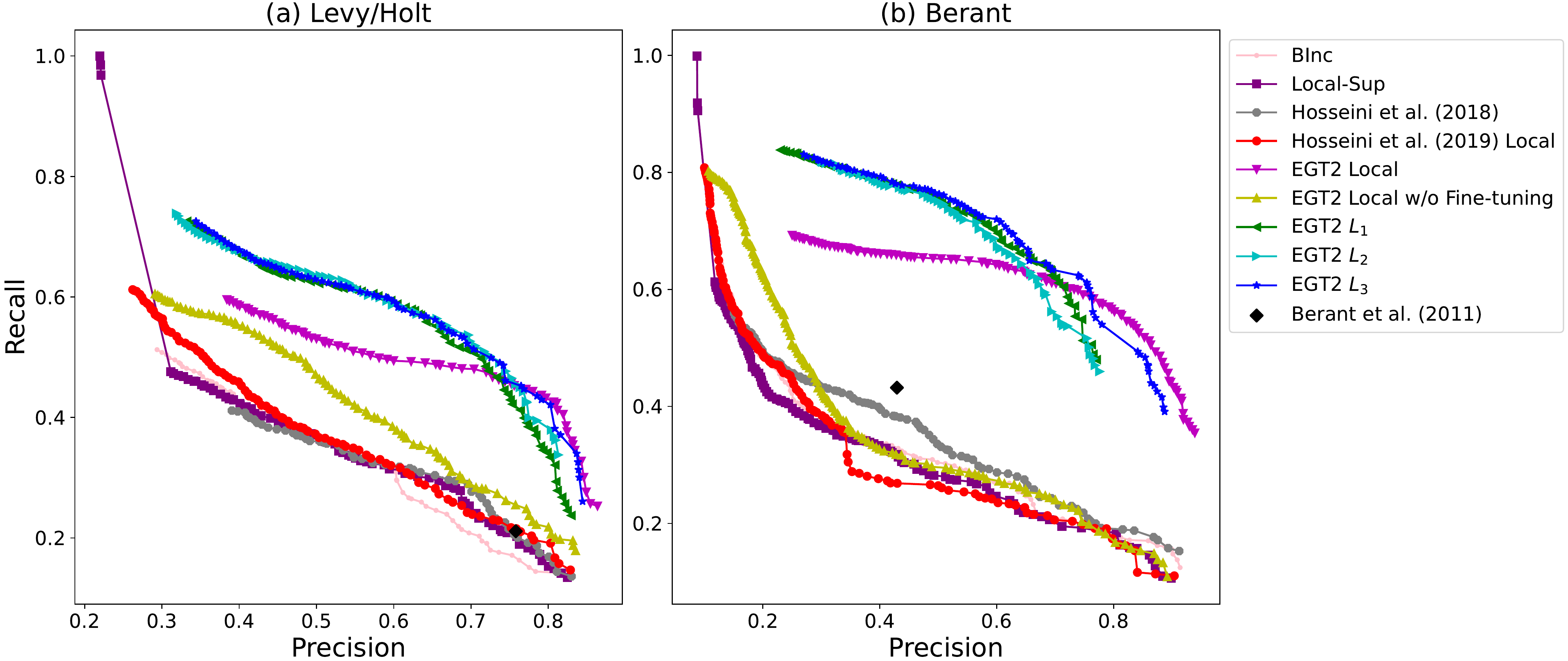}
    \caption{The Precision-Recall Curves of different methods on (a) Levy/Holt Dataset and (b) Berant Dataset. The result of \citet{berant-etal-2011-global} is shown as a point, as they generate entailment graphs without weight.}
    \label{fig:curve}
\end{figure*}

The Precision-Recall Curves of different methods and the Precision-Recall Point of \citet{berant-etal-2011-global} on the two evaluation datasets are shown in Figure \ref{fig:curve}(a) and \ref{fig:curve}(b) respectively. The local and global models of EGT2 consistently outperform previous state-of-the-art methods on all levels of precision and recall, which indicates the effect of our local model based on textual entailment and global soft constraints based on transitivity. The EGT2-Local achieves slightly higher precision than global models in the range $recall<0.5$, but its precision drops quickly if we require higher recall and therefore leads to worse performance than global models. The result indicates that global models with transitivity constraints gain significant improvement on recall with far less expense on precision than EGT2-Local.

\subsection{How the local model fine-tuning works?}
As described in Section~\ref{implement}, a new corpus is generated for fine-tuning the local model. We claim that the fine-tuning corpus helps to improve the performance of EGT2-Local by adapting it to the special sentence pattern by $S$, rather than offering additional data to fit the distribution of target datasets as traditional training datasets do. To prove this, we also test a simple supervised method, labelled as Local-Sup, which fits a 2-layers feedforward neural network on the fine-tuning corpus with cosine similarity, Weed, Lin and BInc scores as features. If the corpus acts as training dataset, the performance of Local-Sup should be obviously better than its unsupervised features.

As shown in Table \ref{result-table}, Local-Sup does not perform significantly better on Levy/Holt Dataset, and even worse on Berant Dataset than BInc, which is one of the inputting features of Local-Sup. The result illustrates the difference between the fine-tuning corpus and the evaluation datasets, and shows that the corpus plays a role as pattern adapting corpus rather than training dataset.

\subsection{Why are global constraints helpful?}

In Section~\ref{intro}, we expect that the improvement of soft transitivity constraints is attributed to the alleviation of data sparsity in corpus. To examine the sparsity before and after the applying of transitivity constraints, we count how many the positive and negative entailment relations in the Levy/Holt test set exactly appear in the local and global entailment graph respectively, and show the counting results in Table \ref{result-count}. All three soft transitivity constraints help to find more entailment relations than local entailment graph and therefore achieve better performance on the evaluation datasets. Although EGT2-$L_2$ finds the most entailment relations in the dataset in global stage, it finds more negative examples concurrently and thus performs worse than $L_1$ and $L_3$ as shown in Table \ref{result-table}. On the other hand, EGT2-$L_1$ and EGT2-$L_3$ obtain more proportions of positive examples by considering premise relations during the gradient calculation. The low confidence of hypothesis relationship $W_{a,c}$ should be helpful to detect spurious premises $W_{a,b}$ and $W_{b,c}$. Therefore, EGT2-$L_3$ slightly outperforms EGT2-$L_1$ as the gradients of $W_{a,b}$ and $W_{b,c}$ in $L_3$ are related to the hypothesis relationship $W_{a,c}$.

\begin{table}
\caption{The number of testing examples appearing in entailment graphs learnt by corresponding models.}\label{result-count}
\centering
\begin{tabular}{l|c|c}
\hline
\textbf{Methods} & \textbf{Positive \#} & \textbf{Negative \#} \\
\hline
EGT2-Local & 378 & 75 \\
EGT2-$L_1$ & 642 & 174 \\
EGT2-$L_2$ & 783 & 277 \\
EGT2-$L_3$ & 685 & 190 \\
\hline
\end{tabular}
\end{table}

We have also applied the soft transitivity constraints on the local graph with BInc and \citet{hosseini-etal-2019-duality}, but observed only slightly improvement of performance, as $.155\to .157$ and $.167\to .170$ for EGT2-$L_3$ on PRC of Levy/Holt Dataset respectively. Comparing it with the significant improvement based on EGT2-Local, we claim that the high-quality local entailment graphs are the basis of effective soft transitivity constraints.

The previous cross-graph soft constraint and paraphrase resolution soft constraint proposed in \citet{hosseini-etal-2018-learning} have shown improvement of performance based on their local graphs. However, due to the distinct distribution and scales of local scores, their constraints are computationally unavailable on our local graphs, partially due to the high overhead for cross-graph calculation.

\subsection{Does EGT2 learn directional entailment?}
\begin{table*}
\caption{Model performance on the directional section of Levy/Holt Dataset with two settings (a) and (b)}.\label{directional-result}
\centering

\begin{tabular}{l|c|c|c|c|c|c}
\hline
\textbf{Setting} & \multicolumn{4}{c|}{\bf (a)} &  \multicolumn{2}{c}{\bf (b)} \\
\hline
\textbf{Methods} & \textbf{valid PRC} & \textbf{valid ROC} & \textbf{test PRC} & \textbf{test ROC} & \textbf{PRC} & \textbf{ROC} \\
\hline
BInc & .456 & .508 & .472 & .509 & .038 & .567 \\
Hosseini18 & .456 & .500 & .488 & .533 & .038 & .564 \\
Hosseini19-Local & .462 & .524 & .506 & .561 & .040 & .579 \\
\hline
EGT2-Local & .500 & .552 & .548 & .606 & .176 & .654 \\
EGT2-$L_3$ & .511 & .596 & .564 & .637 & .171 & .696 \\
\hline
\end{tabular}

\end{table*}

Generally, the logical entailment should be directional which makes it different from \emph{paraphrase}. Although EGT2 significantly improves the performance on two datasets, it is unclear whether the improvement comes from the directional entailment cases, or only paraphrasing ones, as the local LM might be strong in recognizing paraphrases but weak in recognizing directional entailment \citep{cabezudo2020natural}. 

To examine how EGT2 works under directional cases, we use two different settings to generate directional section of Levy/Holt Dataset: (a) remaining those paraphrase predicate pairs $a\to b$ with label $l\in\{True,False\}$ in Levy/Holt validation and test set if the corresponding $b\to a$ also exists but is labelled as $\lnot l$, which is used in previous works~\cite{hosseini-etal-2018-learning} and contains 1,207 positive and 1,207 negative  examples\footnote{\url{https://github.com/mjhosseini/entgraph\_eval/tree/master/LevyHoltDS}}; (b) only eliminating $a\to b$ with label $l$ from Levy/Holt test set if $b\to a$ is also labelled as $l$, and contains 753 positive and 7,387 negative examples\footnote{The main difference between two settings is that if $a\to b$ appears but $b\to a$ does not appear in the dataset, setting (a) will eliminate $a\to b$ from the directional section, while setting (b) remains $a\to b$ in the directional section.}.
We expect that the directional settings could be more challenging as undirectional paraphrases are unavailable in both settings.

We report the model performance on the directional section of Levy/Holt Dataset with two settings in Table \ref{directional-result}. The relatively high scores on AUC of PRC with setting (a) are due to the different positive-negative ratios of the directional section and the original dataset (1:1 and 1:4, respectively). We can see that previous baselines do not perform well on AUC of PRC with setting (b), which indicates that it is difficult for them to reach $precision>0.5$ with the harder setting. Meanwhile, EGT2-Local and EGT2-$L_3$ outperform all baselines on the two directional settings of Levy/Holt Dataset. Unsurprisingly, the AUC scores of all models on the directional section with setting (b) become lower compared on the original Levy/Holt Dataset, showing the challenges of directional entailment inference. Two EGT2 variants maintain high performance, which proves that our local model can learn to capture directional predicate entailment better than distributional baselines, and the global soft constraint also helps to make directional entailment inference.

\subsection{Error Analysis}

We randomly sample and analyze 100 false positive (FP) examples and 100 false negative (FN) examples from Levy/Holt test set according to predictions by EGT2-$L_3$. We manually setup the decision threshold as 0.574 to make the precision level close to 0.76, which is the same as \citet{berant-etal-2011-global}. The major error types are shown in Table \ref{result-erroranaly}. Although the global constraint is used, about half of FN errors are due to the data sparsity where the entailment relations are not found in the entailment graph. When compared with the results in \citet{hosseini-etal-2018-learning}, EGT2-$L_3$ reduces  the ratio of \emph{Sparsity} in FN errors from 93\% to 46\% with stronger alleviation ability of data sparsity. About a quarter of FN are caused by the \emph{Under-weighted Relations} in the graph, where EGT2 finds the entailment relations but gives them scores lower than the threshold. The rest of FN are related to \emph{Dataset Wrong Labels} which happens when the predicates are indeed entailed by others but labelled as negative, or the predicate pairs are incomplete.

Most of FP errors are caused by the \emph{Spurious Correlation} as these relations are too fraudulent for EGT2 to see through their spurious relationships and consequently given high scores. A few FP errors are caused by \emph{Lemma-based Processing} in LM inevitably, but the ratio still reduces from 12\% in \citet{hosseini-etal-2018-learning} to 5\%. The result indicates that our fine-tuned LM can handle the predicates even with similar surface forms and contexts better than parsing-based distributional local features.

\begin{table}
\caption{The major error types of false positive and false negative predictions by EGT2-$L_3$ in Levy/Holt test set, with predicted scores in the parentheses.
}\label{result-erroranaly}
\centering
\begin{tabular}{l|l}
\hline
\textbf{Error Types} & \textbf{Examples} \\
\hline
\emph{False Negative} & \\
\hline
Sparsity (46\%) & \makecell[l]{Pain relieves by application\\of Chloroform. $\to$ Chloro-\\form reduces pain. (0.0)}\\
\hline
\makecell[l]{Under-weighted\\Relations (23\%)} & \makecell[l]{The Druids build the\\Stonehenge. $\to$ The Druids\\construct the Stonehenge.\\(0.558)}\\
\hline
\makecell[l]{Dataset Wrong\\Labels (31\%)} & \makecell[l]{Salicylates reduces pain.\\$\to$ Salicylates is given for\\pain. (0.034)}\\
\hline
\emph{False Positive} & \\
\hline
\makecell[l]{Spurious Cor-\\relation (68\%)} & \makecell[l]{The cat sleeps on a fur. $\to$\\The cat has a fur. (0.683)}\\
\hline
\makecell[l]{Lemma-based\\Process (5\%)} & \makecell[l]{Lincoln comes to New\\York. $\to$ Lincoln comes\\from New York. (0.867)}\\
\hline
\makecell[l]{Dataset Wrong\\Labels (27\%)} & \makecell[l]{The lamps are made of\\metal. $\to$ The lamps are\\made of metal. (1.0)}\\
\hline
\end{tabular}
\end{table}

\section{Conclusions}

In this paper, we propose a novel typed entailment graph learning framework, EGT2, which uses language models fine-tuned on textual entailment tasks to calculate local entailment scores and applies soft transitivity constraints to learn global entailment graphs in gradient-based method. The transitivity constraints are achieved by carefully designed loss functions, and effectively boost the quality of local entailment graphs. 
By using the fine-tuned local LM and global soft constraints, EGT2 does not rely on distributional features, and can be easily applied to large-scale graphs. Experiments on standard benchmark datasets show that EGT2 achieves significantly better performance than existing state-of-the-art entailment graph methods.

\section*{Acknowledgements}

This work is supported in part by National Key R\&D Program of China (No. 2020AAA0106600) and NSFC (62161160339). We would like to thank the anonymous reviewers and action editors for their helpful comments and suggestions.



\bibliography{anthology,custom}
\bibliographystyle{acl_natbib}

\clearpage

\appendix
\section{Algorithm for Sentence Generator}

\label{sec:sg}
\begin{algorithm}[hbt!]
\small
\caption{The sentence generator $S$.} 
\label{alg::sg} 
\begin{algorithmic}[1] 
\Require 
$p=(w_{p,1}.i_{p,1}, w_{p,2}.i_{p,2}, t_1 ,t_2)$: a typed predicate;
\Ensure 
Sentence $S(p)$
\If{Order of ($t_1$,$t_2$) is equal to graph types}
\State{Actor$_1$ = concat($t_1$,"A")}
\State{Actor$_2$ = concat($t_2$,"B")}
\Else
\State{Actor$_1$ = concat($t_1$,"B")}
\State{Actor$_2$ = concat($t_2$,"A")}
\EndIf
\If{The first word of $w_{p,1}$ or $w_{p,2}$ is not a verb}
\State{$w_{p,1}$ = concat("is",$w_{p,1}$)}
\State{$w_{p,2}$ = concat("is",$w_{p,2}$)}
\EndIf
\State{Active$_1$=Boolean($i_{p,1}=1$)}
\State{Active$_2$=Boolean($i_{p,2}=1$)}
\State{MinLen=min(Length($w_{p,1}$),Length($w_{p,2}$))}
\State{MML=max i, s.t.$w_{p,1}$[1:i]=$w_{p,2}$[1:i]}
\State{Pathway=Boolean(MML=MinLen)}
\If{Active$_1$ and Active$_2$}
\If{Pathway}

\Return{concat(Actor$_1$,"and",Actor$_2$,$w_{p,1}$[1:\newline MinLen])}
\EndIf

\Return{concat(Actor$_1$,"and",Actor$_2$,$w_{p,1}$[1])}
\EndIf
\If{Active$_1$ and not Active$_2$}
\If{Pathway}
\State{Act=$w_{p,1}$}
\If{Length($w_{p,1}$)<MinLen}
\State{Act=$w_{p,2}$}
\EndIf
\Return{concat(Actor$_1$,Act,Actor$_2$)}
\EndIf
\Return{concat(Actor$_1$,$w_{p,1}$,"Something",\newline$w_{p,2}$[MML+1:],Actor$_2$)}
\EndIf
\If{Active$_2$ and not Active$_1$}
\If{The first words of $w_{p,1}$ is verb}
\Return{concat(Actor$_1$,Reverse(\newline$w_{p,2}$[MML:]),"to",$w_{p,1}$,Actor$_2$)}
\EndIf
\Return{concat(Actor$_1$,Reverse($w_{p,2}$),\newline$w_{p,1}$[MML:],Actor$_2$)}
\EndIf
\If{Pathway}
\Return{concat(Actor$_1$,Passive($w_{p,1}$),\newline$w_{p,2}$[MML:],Actor$_2$)}
\EndIf
\Return{concat("Something",$w_{p,1}$,Actor$_1$,\newline$w_{p,2}$[MML:],Actor$_2$)}
\end{algorithmic} 
\end{algorithm}

\newpage
\section{Additional Implementation Details}
\label{addi-detail}

We select the SGD learning rate $\alpha$ from $\{0.02,0.05,0.1\}$, the number of training epochs from $\{2,3,5,7\}$, the coefficient $\lambda$ from $\{0.5,1,2\}$, and the confidence threshold $\epsilon$ from $\{0.005,0.01,0.02\}$. We manually tune the hyper-parameters based on the AUC of PRC on Levy/Holt validation dataset, which is .327 corresponding to our settings. 

Under our experiment settings, one training epoch costs about 4 hours on an NVIDIA A40 GPU.

\end{document}